# Adversarial vs behavioural-based defensive AI with joint, continual and active learning: automated evaluation of robustness to deception, poisoning and concept drift

Alexandre Dey[2], Marc Velay[1], Jean-Philippe Fauvelle[1], Sylvain Navers[1]

[1] Airbus Defence & Space Cybersecurity, 1 boulevard Jean Moulin 78990 Élancourt
[2] Airbus Defence & Space Cybersecurity, 3 rue Louis Braille 35136 Saint-Jacques-de-la-Lande

Contact: jean-philippe.fauvelle@airbus.com

**Abstract:** Recent advancements in Artificial Intelligence (AI) have brought new capabilities to behavioural analysis (UEBA) for cyber-security consisting in the detection of hostile action based on the unusual nature of events observed on the Information System. In our previous work (presented at C&ESAR 2018 and FIC 2019), we have associated deep neural networks auto-encoders for anomaly detection and graph-based events correlation to address major limitations in UEBA systems. This resulted in reduced false positive and false negative rates, improved alert explainability, while maintaining real-time performances and scalability. However, we did not address the natural evolution of behaviours through time, also known as concept drift. To maintain effective detection capabilities, an anomaly-based detection system must be continually trained, which opens a door to an adversary that can conduct the so-called "frog-boiling" attack by progressively distilling unnoticed attack traces inside the behavioural models until the complete attack is considered normal. In this paper, we present a solution to effectively mitigate this attack by improving the detection process and efficiently leveraging human expertise. We also present preliminary work on adversarial AI conducting deception attack, which, in term, will be used to help assess and improve the defense system. These defensive and offensive AI implement joint, continual and active learning, in a step that is necessary in assessing, validating and certifying AI-based defensive solutions.

# 1    Introduction

Les systèmes actuels de surveillance de la sécurité des Systèmes d'Information (SI) reposent sur des règles statiques de détection, conçues par capitalisation et maintenues postérieurement à l'observation de nouvelles cyber-attaques. L'évolution incessante et la complexité croissante des menaces surchargent les différents niveaux d'expertise opérant ces systèmes. Cette expertise rare devrait être focalisée davantage sur des tâches à haute valeur ajoutée plutôt que sur des tâches répétitives chronophages.

De nombreux travaux récents consacrés à la détection d'intrusion dans un SI, reposent sur l'analyse du comportement des utilisateurs et des entités ou « UEBA » (*User and Entity Behaviour Analytics*). Le procédé UEBA le plus souvent mis en œuvre repose sur des algorithmes d'apprentissage machine pour construire des modèles de comportements normaux puis inférer, pour chaque évènement observé, un score d'incongruité qui quantifie la déviance dudit comportement observé relativement au modèle appris. Cette approche avantageuse rend possible la détection d'attaques non encore observées et permet une adaptation graduelle aux évolutions du SI protégé.

Cependant, les solutions existantes qui mettent en avant cette approche UEBA concèdent certaines limitations, principalement un nombre élevé de faux positifs, une sur-simplification des problèmes entraînant des faux négatifs, un passage à l'échelle difficile dans certains cas, un effet « boîte noire » inhérent au caractère faiblement explicable des alertes, et enfin une adaptabilité insuffisante à l'évolution naturelle des comportements dans le temps ou « dérive conceptuelle » (*concept drift*).

Dans nos précédents travaux [1] nous avons présenté des avancées associant un procédé UEBA avec une phase de corrélation d'évènements organisés en graphes, permettant de nous affranchir des limites susmentionnées, à l'exception de la dérive conceptuelle.

Nous contournons cette dérive conceptuelle en recourant à un <u>apprentissage continu</u>, mais ceci a pour conséquence d'exposer notre procédé de détection à une nouvelle menace : un attaquant pourrait tromper un système de détection apprenant en continu [2, 3, 4] en disséminant progressivement les traces de son attaque jusqu'à ce que cette dernière soit considérée comme normale par l'outil de détection [5, 6]. Ce type d'attaque par empoisonnement se nomme « grenouille ébouillantée » (*frog-boiling*) [7]. Nous proposons d'adresser cette problématique de la grenouille ébouillantée par la mise en œuvre d'un <u>apprentissage actif</u> (*active learning*) (Table 1).

En outre, notre procédé défensif est exposé à une menace commune à tout système de détection d'anomalies, consistant à leurrer la défense en menant une activité discrète lors de la phase offensive : un attaquant qui aurait connaissance du mécanisme de défense mis en œuvre sur le SI visé pourrait imiter un comportement normal afin de leurrer les défenses. C'est notamment l'approche pratiquée par les APT (*Advanced Persistent Threat*) pour créer des attaques complexes. Ce type d'attaque se nomme « attaque adversaire » (*adversarial attack*) et peut être automatisé via un apprentissage par renforcement [27, 28]. Nous proposons d'adresser cette problématique de l'attaque adversaire par la mise en œuvre d'un <u>apprentissage joint</u> adversaire versus défensif (Table 1).

En outre, cet apprentissage joint a également pour effet de compenser en partie le manque récurrent de données d'entrainement (*training data*), massives, cohérentes et labélisées qui demeurent indispensables à la validation du fonctionnement du procédé (Table 1).

En prolongement de nos travaux exposés lors de la Conférence C&ESAR 2018 [21] puis du FIC 2019, nous présentons dans cet article, un système défensif de détection de cyber-attaques fonctionnant en <u>apprentissage continu</u> et <u>apprentissage actif</u>, ainsi qu'un système d'attaque adversaire couplé au système défensif via un <u>apprentissage joint</u>, le tout permettant, d'une part, d'adresser l'ensemble des problématiques sus-évoquées, et d'autre part, d'évaluer notre propre système défensif dans un premier temps puis d'autres systèmes défensifs par la suite.

**Table 1.** — Les procédés qui résolvent ou provoquent les problématiques.

|  |  | **Problématiques** | | | |
|---|---|---|---|---|---|
|  |  | Données d'entrainement | Attaque adversaire | Dérive conceptuelle | Grenouille ébouillantée |
| **Procédés** | Apprentissage joint | Résout (partiellement) | Résout |  |  |
|  | Apprentissage continu |  |  | Résout | Provoque |
|  | Apprentissage actif |  |  |  | Résout |

## 2   Principes généraux

Le présent chapitre introduit les principes nécessaires à la compréhension de l'article.

**Procédé UEBA et dérive conceptuelle.**

Un procédé UEBA comprend généralement une phase d'apprentissage construisant un modèle des comportements habituels des constituants d'un SI, ainsi qu'une phase d'inférence comparant chaque comportement au modèle pour en mesurer le caractère inhabituel c'est-à-dire potentiellement hostile.

Le plus souvent, la phase d'apprentissage est menée une fois pour toute, puis le modèle demeure statique : un comportement pourra alors être perçu normal ou hostile, à tort ou à raison selon l'obsolescence du modèle inhérente à la dérive conceptuelle.

Ce phénomène de dérive conceptuelle peut être adressé en instaurant un apprentissage continu, mais ceci expose alors le système aux attaques par empoisonnement [33], où un attaquant mène des actions d'apparence normale qui tendent progressive-

Page **3** sur **25**

ment vers l'attaque complète, jusqu'à ce que cette dernière soit intégrée dans les modèles comportementaux et ne puisse plus être détectée.

**Leurrage d'une IA.**

Le leurrage d'un système de détection d'anomalies consiste à dissimuler le caractère hostile d'un ensemble d'actions en mimant un comportement normal. L'approche la plus commune consiste à privilégier des actions considérées comme des signaux faibles et entre lesquels il est difficile d'établir une corrélation.

**Ethical hacking.**

Le *ethical hacking* est une pratique repandue dans le domaine de la cyber-sécurité, prenant la forme d'un jeu opposant deux équipes d'experts, l'une nommée équipe bleue (*blue team*) en charge de défendre un SI, et l'autre nommée équipe rouge (*red team*) en charge de l'attaquer, sans occasionner aucun dommage réel.

**Cyber Kill Chain.**

Ce concept, largement accepté par de nombreux acteurs gouvernementaux et industriels, divise une cyber-attaque en sept phases successives dont chacune comprend ses propres enjeux et contraintes et peut être mitigée par des défenses spécifiques [36].

**IA adversaire.**

Par généralisation et automatisation du *ethical hacking*, une IA adversaire est opposée à une équipe défensive (humaine et/ou IA), dont elle tente de cerner et contourner les limites [22], ce principe étant applicable à des systèmes logiques ou physiques [29].

Il existe principalement trois approches adversaires :

— La première approche repose sur l'apprentissage en parallèle des modèles offensifs et défensifs [32], chacun ayant accès à des informations sur son opposant, qui lui permettent d'améliorer sa propre approche. Leurs optimisations dépendent de la valeur de l'erreur de classification du modèle défensif : ce dernier est entrainé à minimiser l'erreur, tandis que l'adversaire cherche à la maximiser ;
— Une seconde approche consiste à opposer le modèle adversaire à une défense statique dont les classifications sont connues [27] [28] : le problème est résolu en apprenant par renforcement, en fonction de récompenses dont les valeurs sont liées à la réussite ou à l'échec de chaque action ;
— En troisième approche, le modèle adversaire est opposé à un environnement complexe simulé : l'apprentissage ne dépend pas d'un retour direct du système défensif mais d'une mesure de l'effet des actions de l'adversaire sur l'environnement. Nous avons suivi cette approche adversaire dans nos travaux.

# 3 Approche défensive (*blue team*)

Notre procédé défensif dans son état actuel (Fig. 1) comprend trois couches, de détection (pastille 10), de corrélation (pastille 20) et d'apprentissage actif homme-machine basé sur une rétroaction (pastille 30). Le présent chapitre expose les tenants et aboutissants ayant conduit à cet état.

**Fig. 1.** — Synoptique du procédé défensif.

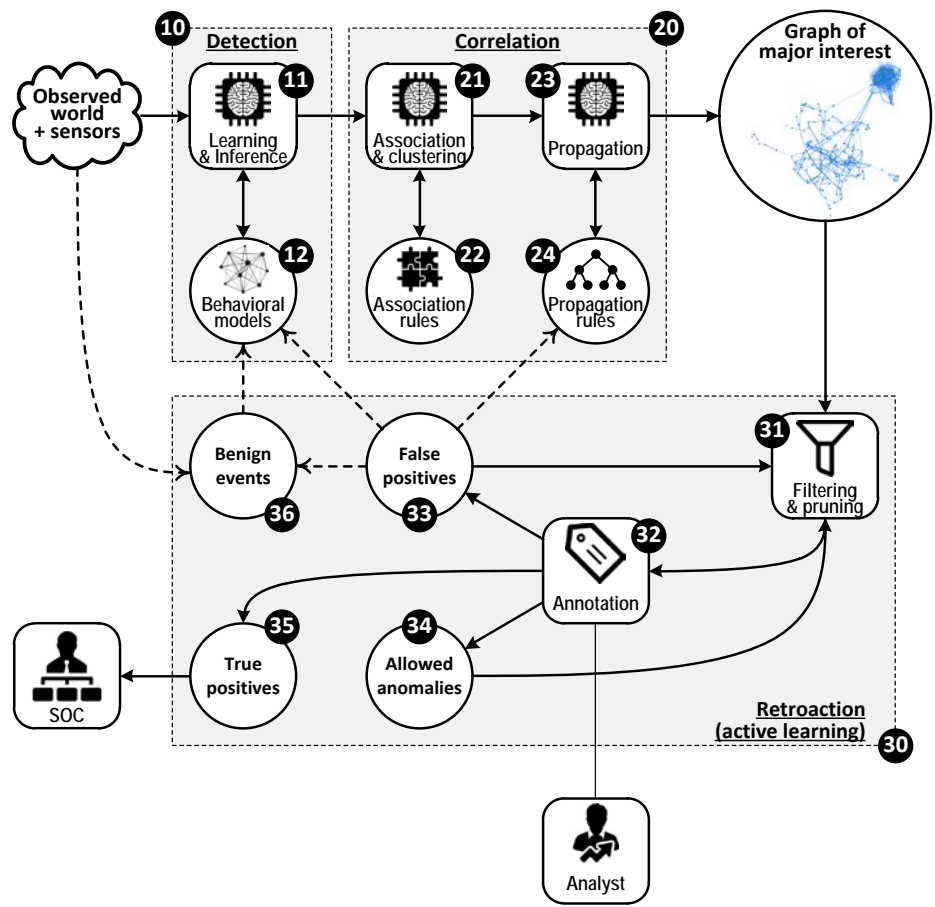

## 3.1 Objectifs

Notre recherche d'un procédé défensif est principalement guidée par ces objectifs :

— Minimiser le nombre de faux positifs par rapport aux solutions existantes ;
— Maximiser le caractère explicable des alertes remontées à l'analyste, c'est-à-dire fournir à ce dernier les informations lui permettant de comprendre les diffé-rentes étapes de l'attaque ainsi que la nature et la force de ce qui les relie avec certitude ou probabilité ;
— Minimiser le temps et l'expertise qui sont nécessaires au déploiement et au fonctionnement du procédé ;
— Maximiser la robustesse du procédé défensif aux attaques par empoisonnement, afin de contrevenir aux actions d'un attaquant. À titre d'exemple, [18] propose une variante robuste de l'algorithme PCA (*Principal Component Analysis*).

### 3.2   Couche de détection

Notre couche de détection (Fig. 1 : pastille 10) basée sur des réseaux de neurones auto-encodeurs (pastille 11) génère et maintient un modèle des comportements (pastille 12) et enrichit les évènements observés en leur adjoignant un score d'incongruité qui quantifie l'écart entre chaque évènement observé et le modèle appris.

Collecter et maintenir une base de données des attaques visant spécifiquement un système à surveiller est particulièrement difficile. La méthode qui semble la plus viable pour servir cet objectif repose sur la détection d'anomalies comportementales. Dans cette voie, les deux exigences majeures sont l'adaptation aux différents types de variables (ex. : nombres, catégories, séries temporelles) qui existent dans les événements de sécurité (ci-après : exigence « **EX-1** ») et la production d'alertes pourvues d'un caractère explicable (ci-après : exigence « **EX-2** »).

Les réseaux de neurones auto-encodeurs sont utilisables pour générer des plongements (*embedding*) [34] pour un grand nombre de formats de données, ce qui répond à la première exigence **EX-1**.

Les auto-encodeurs (Fig. 1 : pastille 11) sont entrainés à reproduire les données d'entrée après une phase de compression. Les anomalies deviennent détectables en raison d'une forte erreur de reconstruction des données par l'auto-encodeur. Cette erreur de reconstruction est particulièrement élevée pour les variables responsables de l'anomalie (ex. : partie correspondant à l'exploitation dans une requête web malveillante). L'algorithme répond donc également à la seconde exigence **EX-2**.

De plus, la capacité de passage à l'échelle des réseaux de neurones rend ce procédé intéressant pour du calcul au fil de l'eau. Ces calculs peuvent également être réalisés sur des architectures matérielles spécialisées (ex. : GPU, TPU, FPGA, ASIC). Cependant, l'entraînement d'un auto-encodeur nécessite un large corpus de données.

Notre contexte d'apprentissage étant soumis au risque de sur-apprentissage du fait des différentes problématique liées à la labellisation (ex. : biais ou erreur de labellisation, obsolescence des données), nous nous appuyons sur les différentes méthodes évoquées dans l'état de l'art afin de réduire ce risque. Outre l'emploi de méthodes de régularisation, pendant l'entraînement, nous contrôlons régulièrement les performances des modèles sur un jeu de données de validation et nous interrompons l'apprentissage de manière anticipée lorsque les performances se dégradent.

### 3.3 Couche de corrélation

Notre couche de corrélation (Fig. 1 : pastille 20) basée sur des algorithmes de graphes, filtre les évènements et les regroupe par liens de causalité.

Un attaquant conduit son opération hostile en plusieurs étapes, produisant une multitude d'évènements. Une anomalie isolée ne peut être considérée à elle seule comme une attaque, car l'ajout d'un score d'incongruité par le procédé UEBA à chaque évènement capté n'est pas suffisant pour détecter une attaque ni a fortiori la reconstruire.

Pour cette raison, notre approche pour détecter et reconstruire de manière explicable une opération hostile s'appuie sur la corrélation d'un ensemble d'évènements représentés sous la forme d'un graphe, cette technique étant communément considérée comme la plus performante selon l'état de l'art [12].

**Compression des structures fréquentes.**

Du fait de la diversité des capteurs et de la complexité des SI, un unique phénomène peut déclencher une cascade d'évènements concomitants (ex. : démarrage d'une application suivi invariablement du chargement de ses dépendances), ou des évènements répétés similaires (ex. : chargement d'un fichier bloc par bloc).

Cette abondance d'évènements augmente la taille des graphes et par conséquent la complexité algorithmique spatiale et temporelle des procédés qui les manipulent.

Nous adressons cette problématique en regroupant les évènements (Fig. 1 : pastille 21), soit par mise en œuvre de règles d'association (pastille 22) dans le cas des évènements concomitants, soit par *clustering* réalisé par un réseau de neurones auto-encodeurs dans le cas des évènements répétés similaires.

**Construction d'un graphe.**

Chaque évènement capté sur un SI est porteur à minima des informations relatives à l'acteur (ex. : processus, machines), à l'action (ex. : lecture, écriture, appel système), au moment (horodatage), à la localisation (ex. : site, équipement, application) et parfois aux ressources sollicitées (ex. : fichier).

À partir de ces éléments, nous construisons des graphes temporels acycliques orientés, constitués de nœuds (ex. : acteurs, ressources) reliés par des arêtes unidirectionnelles (actions), cette représentation étant appropriée pour modéliser un flux d'informations liées entre elles (Fig. 1 : pastille 23).

**Propagation du niveau d'anomalie.**

Le recours aux réseaux de neurones auto-encodeurs nous procure une quantification du niveau d'anomalie de chacune des variables des événements observés.

Cependant, certaines actions sont plus susceptibles que d'autres de caractériser des anomalies sur un type de variable (ex. : un navigateur et un interpréteur de commandes communiquent parfois vers des adresses IP inconnues, mais le navigateur est plus fréquemment impliqué).

Nous adressons cette problématique en favorisant les transferts rares et en pénalisant ceux plus fréquents afin de privilégier les liens incongrus, lors de la création des modèles de propagation du niveau d'anomalie (Fig. 1 : pastille 24).

### 3.4 Couche de rétroaction

Notre couche de rétroaction (Fig. 1 : pastille 30) implémente un apprentissage actif en permettant à un analyste de transmettre progressivement son savoir à notre procédé défensif.

Notre méthode repose sur différents modèles construits par des algorithmes d'apprentissage à partir des données collectées sur le SI. Ces algorithmes sont efficaces pour détecter des anomalies mais non pour les interpréter, cette tâche étant dévolue à un analyste, pour lequel il convient de simplifier les représentations et d'abstraire le fonctionnement sous-jacent du procédé.

**Présentation à l'analyste.**

En présence d'un graphe dit d'intérêt majeur [1], constitué d'événements liés, et dont le niveau d'anomalie est élevé[1], nous élaguons (Fig. 1 : pastille 31) les chemins les moins intéressants i.e. dont le niveau d'anomalie ne fait que décroître. Ce graphe, qui correspond à une alerte présumée, est alors comparé aux anciennes alertes statuées faux positifs regroupées dans une base (pastille 33). En présence de similitudes notables, le graphe d'intérêt majeur est abandonné et aucune alerte n'est émise. Dans le cas contraire, le graphe est présenté à l'analyste, afin de faciliter ses investigations, en même temps que les traces liées aux évènements.

**Sélection des évènements.**

L'analyste porte la responsabilité d'annoter (Fig. 1 : pastille 32) chaque alerte :

— En présence d'un vrai positif (pastille 35) une alerte est envoyée au SOC (*Security Operations Center*) ;
— En présence d'un faux positif, les évènements sont insérés dans une base des faux positifs (pastille 33) destinée au filtrage ;
— En présence d'un comportement dangereux mais autorisé (ex. : actions d'un administrateur), ce dernier est répertorié dans une base des anomalies autorisées (pastille 34).

---
[1] Les modalités décidant qu'un graphe est d'intérêt majeur ne peuvent pas être divulguées.

Lorsque des faux positifs similaires se multiplient, ils contribuent à l'apprentissage.
En outre, notre procédé comprend une base de connaissances des événements bénins passés (pastille 36) c'est-à-dire des données normales destinées à l'entrainement. Cette base fait office de mémoire épisodique (i.e. rejouer des évènements passés marquants pour éviter leur oubli trop rapide) ou permet de reconstruire les modèles en cas de nécessité. Elle est alimentée par échantillonnage aléatoire sur le flux continu des évènements normaux (i.e. ayant un faible score d'anomalie). Elle est également enrichie par les faux positifs que l'on souhaite réintroduire dans les modèles.

Une fois mis au point le procédé défensif, nous étudierons la possibilité d'annoter automatiquement les alertes afin d'alléger le travail de l'analyste.

## 4    Approche offensive (*red team*)

Notre procédé offensif dans son état actuel, présenté dans le présent chapitre, associe une majorité d'actions manuelles et une minorité d'actions automatisées, ces dernières étant menées par une IA adversaire. Cette IA adversaire, qui permet de valider le fonctionnement de l'IA défensive, mime l'approche courante d'un test d'intrusion, hors étape d'ingénierie sociale.

### 4.1    Interaction avec l'environnement

L'IA adversaire est positionnée comme un acteur externe qui accéderait aux ressources internes du SI ciblé puis tenterait d'identifier des vulnérabilités.

Afin de conserver le caractère générique du procédé, l'interfaçage avec l'environnement est réalisé via un terminal et des *sockets* réseau. Pour manipuler cet environnement, l'IA adversaire sélectionne soit le programme et ses options, soit le protocole et le contenu à transmettre via le réseau. L'espace des combinaisons d'actions étant trop vaste à explorer, nous recourons à un dictionnaire de commandes de bas niveau afin d'en réduire la profondeur. Cette faible granularité permet à l'IA adversaire de générer une importante diversité de combinaisons d'actions dont les signatures s'éloignent de celles produites par les outils populaires de cyber-sécurité et demeurent ainsi moins aisément détectables.

Pour chacune des sept phases constituant la *cyber kill chain*, un agent adversaire spécialisé pour ladite phase peut être entrainé à partir d'un dictionnaire spécifique.

À ce jour, notre procédé est encore incomplet :

— Seules les phases 1 et 6 (Reconnaissance ; Commande & Contrôle) de la *Cyber Kill Chain* sont partiellement automatisées, les autres phases demeurant pour l'instant manuelles (Table 2) ;
— L'orchestration des agents reste séquentielle, cette limite ayant toutefois un avantage : un agent est en mesure d'utiliser les informations produites par les agents exécutés antérieurement.

Nous réfléchissons à la mise en œuvre simultanée et collaborative de l'ensemble des agents pour l'ensemble des phases de la *Cyber Kill Chain*.

### 4.2  Scénarios de validation

Afin de valider le fonctionnement des défenses face à diverses attaques, nous avons conçu trois scénarios présentés ci-après, dans lesquels les agents adversaires sont entrainés conformément à la troisième approche adversaire précédemment exposée dans la section « **IA adversaire.** » du chapitre §2 :

— Le premier scénario permet de valider la résistance à une attaque classique. Il prend place dans un SI simulé comprenant plusieurs zones, notamment une zone « utilisateurs » constituée d'ordinateurs bureautiques, une zone « applicatifs » constituée de serveurs hébergeant les applications de l'entreprise, et une zone « SOC ». Ces zones sont équipées de sondes réseau collectant des informations sur les communications et les transmettant aux agrégateurs de la zone SOC. Dans un contexte d'apprentissage machine, ce SI générique est nommé « environnement ». Notre système défensif est installé dans la zone SOC, mais n'est pas activé. En effet, l'intérêt de ce scénario est de déterminer si des éléments internes sont exposés par erreur et si des vulnérabilités courantes sont présentes. Ce scénario est l'équivalent d'un test d'intrusion classique, superficiel, mais nécessaire pour rapidement découvrir des failles évidentes ;

— Le second scénario a pour objectif de valider le bon fonctionnement du système de détection d'attaque. Notre système défensif, préalablement entrainé sur une activité normale générée synthétiquement, mais n'apprenant plus, est actif et analyse le comportement des utilisateurs et entités du SI. Dans ce contexte, l'IA adversaire essaye de leurrer les défenses afin de ne pas être détectée pendant sa découverte et son exploitation du SI ;

— Le troisième scénario entend valider la résistance à l'empoisonnement de la mémoire de notre système défensif, actif et apprenant en continu. L'IA adversaire essaye de corrompre graduellement la représentation d'un comportement normal.

Dans ces trois scénarios, l'IA adversaire explore différentes stratégies. L'exploration est une phase d'adaptation par renforcement où l'agent détermine le moyen le plus efficace d'attaquer le SI. Une méthode heuristique est utilisée pour amener l'IA adversaire à analyser la plus grande surface possible de l'environnement, en termes de cibles et de techniques offensives. L'efficacité des agents adversaires est mesurée via le taux de couverture (surface d'attaque explorée rapportée à la surface existante) et l'efficacité (quantité d'actions pour réaliser une attaque).

### 4.3 Évaluation de l'environnement

L'apprentissage des agents adversaires nécessite une quantification de la réponse de l'environnement aux actions réalisées. De plus, la planification des prochaines étapes de l'opération nécessite l'extraction d'informations dans un format que les algorithmes peuvent exploiter.

Ces deux tâches sont réalisées par analyse des réponses aux requêtes. Chaque requête possède un format spécifique de sortie dont nous pouvons tirer parti pour extraire des informations, par exemple les adresses IP et les ports ouverts des cibles, les versions des logiciels ou encore la confirmation de lancement d'un terminal. Ces premières informations brutes permettent aux modèles de sélectionner l'action suivante la plus appropriée. Elles permettent également d'évaluer le fonctionnement des agents selon plusieurs critères.

Ces critères sont définis en fonction de l'objectif d'apprentissage désiré pour la phase. À titre exemple pour la phase d'identification des vulnérabilités, le taux de couverture de l'analyse et sa discrétion sont les deux critères majeurs. Le taux de couverture est obtenu en comparant les informations découvertes par rapport aux informations connues relatives à une infrastructure maîtrisée.

L'évaluation produit un score que l'algorithme d'apprentissage par renforcement cherche à maximiser. Un certain nombre d'algorithmes fonctionnent sur un score normalisé entre 0 et 1, mais d'autres ne sont pas limités par une valeur maximale.

### 4.4 Apprentissage par renforcement

L'algorithme d'apprentissage par renforcement de tâches complexes A3C (*Asynchronous Actor-Critic Agent*) permet d'explorer un espace d'actions continu ou discret. Sa structure reprend l'approche des GAN [32] en combinant deux structures disjointes.

Dans cette approche :

— L'agent acteur choisit une action suivant une politique dépendante de l'environnement ;
— L'agent critique analyse l'effet des actions entreprises sur l'environnement et calcule un score associé ;
— L'acteur met à jour sa politique d'action pour améliorer ce score.

L'apprentissage est une itération d'essais, d'analyse et d'adaptation (Fig. 2) :

— Un agent « acteur » (Fig. 2 : pastille 1) construit une commande à partir du dictionnaire à disposition, puis l'exécute sur le SI ciblé (pastille 2) ;
— Le message en sortie de cette commande est analysé (pastille 3) et les informations en sont extraites automatiquement. Ces informations représentent l'état actuel de l'environnement ;
— L'agent « critique » (pastille 4) utilise les données analysées précédemment pour évaluer l'acteur ;

— En prenant en compte le nouvel état de l'environnement et son score, l'acteur modifie sa politique de génération de commandes pour l'améliorer, puis le cycle recommence pour continuer l'opération.

Des répliques du modèle, dans des environnements identiques mais séparés, peuvent être exécutées simultanément. Chaque acteur réalise une suite d'actions jusqu'à atteindre une condition d'arrêt. L'apprentissage peut être réalisé par lots (*batch*), en combinant des séquences d'actions. L'algorithme combine les différentes expériences des modèles afin d'en tirer des leçons plus générales. Cette approche permet un passage à l'échelle horizontal des agents.

L'algorithme mis en œuvre par les agents se trouve confronté à un dilemme permanent : soit favoriser l'exploitation d'une stratégie qui procure un score élevé, soit favoriser l'exploration de nouvelles actions. Un déséquilibre entre ces deux voies peut engendrer un sur-apprentissage lorsque l'exploitation est excessivement favorisée au détriment de l'exploration, ou une carence en apprentissage dans le cas opposé. Nous limitons ces risques en parallélisant des agents exposés à des environnements et initialisations différents, qui produisent des suites d'actions différentes, pour entrainer un modèle racine dont sont tirés les agents à chaque épisode. Par ailleurs, nous introduisons un coefficient pendant l'inférence qui induit l'exécution occasionnelle d'actions jugées non optimales afin de provoquer des explorations.

**Fig. 2.** — Synoptique du procédé A3C d'apprentissage par renforcement des agents.

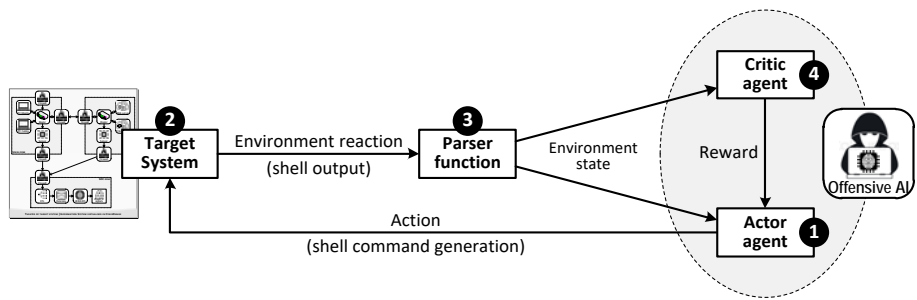

### 4.5 Apprentissage joint

L'apprentissage joint (*joint learning*) consiste à asservir autour d'un SI l'entraînement des IA défensive et adversaire : l'adversaire exécute des actions sur le SI, produisant des évènements collectés qui alimentent l'IA défensive ; cette dernière reconstruit des graphes d'actions, dotées d'un score d'incongruité, en corrélant des événements.

Lors de l'apprentissage, les IA se transmettent leurs sorties respectives : l'adversaire transmet au défenseur son graphe d'actions (*attack graph*) tandis que l'IA défensive transmet son score d'incongruité (*threat score*) à l'adversaire (Fig. 3).

La finalité d'un tel procédé est de réduire l'erreur de reconstruction des graphes et minimiser le score d'incongruité. En théorie, cette approche permet à l'adversaire de développer une stratégie de leurrage dont l'effet se mesure via la dégradation du score de détection au fil de l'attaque. Dans le cas d'un leurrage réussi, l'analyse des actions effectuées par l'agent offensif permet d'identifier les angles morts du système de défense afin de les corriger.

**Fig. 3.** — Synoptique du procédé d'apprentissage joint offensif versus défensif.

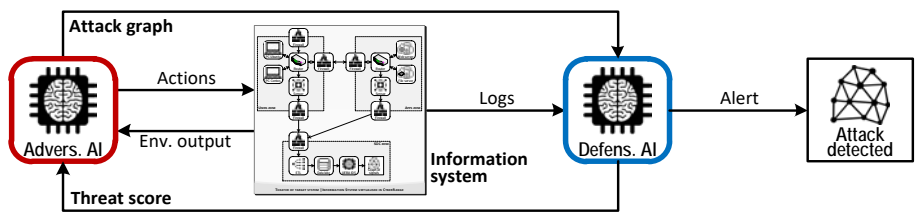

## 5  Expérimentation

Nous avons soumis notre approche à une preuve de concept dont le présent chapitre expose les grandes lignes.

### 5.1  Théâtre

Le théâtre accueillant nos procédés de défense et d'attaque, constituées d'éléments virtualisés sur une plateforme CyberRange[2], est représentatif du SI commun moyennement protégé d'une entreprise de taille moyenne (Fig. 4 ; Fig. 5).

Ce système d'information simulé est ainsi compartimenté :

- La zone 1 (en DMZ) dévolue aux échanges entre le SI de l'entreprise et l'extérieur comprend notamment un *proxy* et un *reverse-proxy* (*web proxy / r-proxy* sur la figure) destinés respectivement aux flux sortant et entrant du SI ;
- La zone 2 contient notamment un site de e-commerce hébergé sur un serveur Web (*web server* sur la figure) ;
- La zone 3 contient les postes de travail (PC) des utilisateurs (employés) ;
- La zone 4 contient les services d'infrastructure et en particulier un serveur de fichiers (*file server* sur la figure) ;
- La zone 5 est dévolue à l'administration du SI ;
- La zone 6 correspond à un SOC d'entreprise équipé d'un outil SIEM (*Security Information and Event Management*). En outre, cette zone contient notre procédé défensif puisque telle serait sa place sur un réel SI d'entreprise.

Notre procédé adversaire fait partie du théâtre d'expérimentation mais il est évidemment positionné hors du SI simulé, auquel il accède via un réseau Internet également simulé.

---
[2] Solution de modélisation et de simulation avancées, conçue par Airbus CyberSecurity.

Nous employons un logiciel d'émulation d'entrées/sorties clavier et souris qui simule des utilisateurs inter-opérant avec le SI, afin de générer un trafic de vie massif et réaliste que nous utilisons pour la mise en point de l'IA défensive.

Les principales caractéristiques de ce trafic de vie sont les suivantes :

— Chaque évènement comprend généralement de 5 à 15 métriques ;
— Nous produisons environ 100 évènements par seconde par machine, ce qui correspond approximativement à 40 millions de métriques par heure pour l'ensemble du SI ;
— La ventilation de ce trafic en fonction de sa source est d'environ 60% pour le système d'exploitation, 30% pour le réseau et 10% pour les applications ;
— La ventilation de ce trafic en fonction de son usage est d'environ 60% pour l'entrainement de l'IA, 20% pour la validation et 20% pour les tests.

**Fig. 4.** — Vue simplifiée du théâtre simulé. Le procédé défensif est intégré au SOC, tandis que le procédé offensif est positionné hors du SI, accédant à ce dernier via Internet.

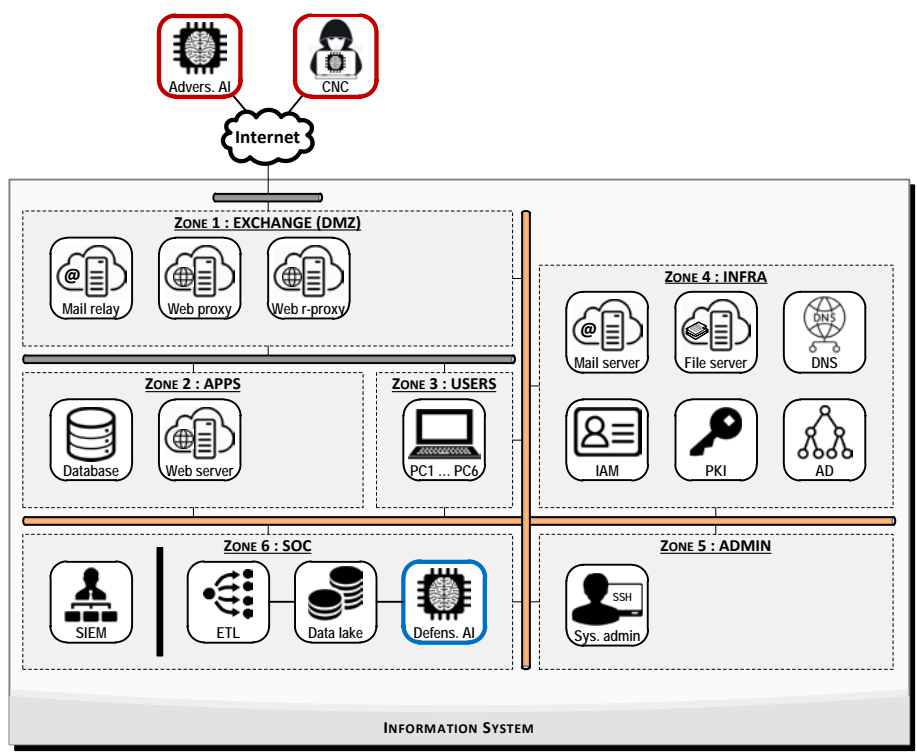

**Fig. 5.** — Vue technique du théâtre simulé (machines virtuelles).

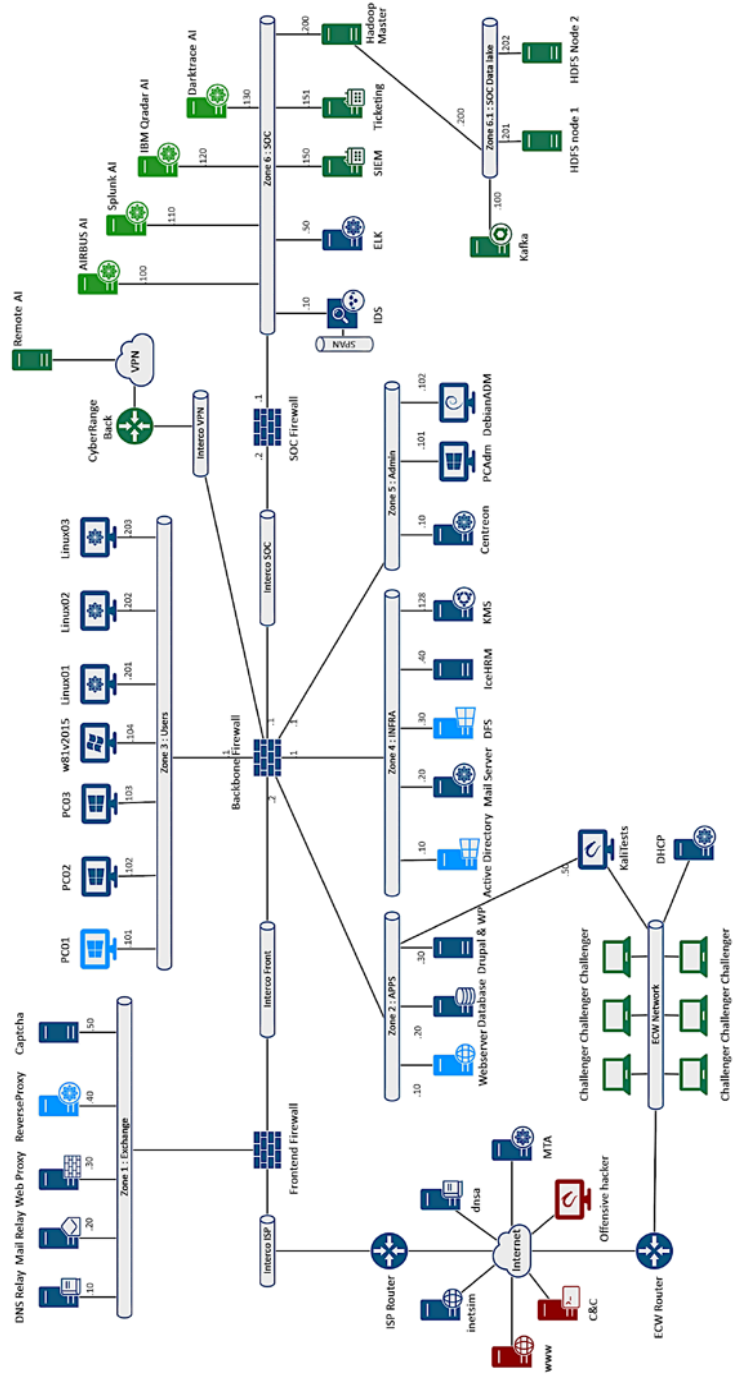

## 5.2 Scénario d'attaque

Afin, d'une part, d'évaluer les capacités de détection de l'IA défensive confrontée à l'IA adversaire, et d'autre part, de s'assurer qu'un faux négatif serait imputable à une insuffisance de l'adversaire plus qu'à un manquement du défenseur, nous avons conçu un scénario d'attaque qui se conforme au modèle de la *Cyber Kill Chain* (Table 2 ; Fig. 6 ; Fig. 7).

**Table 2.** — Étapes du scénario d'attaque.

| Phase de la *Cyber Kill Chain* | Adversaire | Action du scénario d'attaque |
|---|---|---|
| **Phase 1** Reconnaissance | IA | **Phase 1a** : l'IA adversaire explore le réseau (scan de ports). |
| | Humain | **Phase 1b** : l'attaquant découvre le site de e-commerce contenant une fonction de vérification de la version du navigateur. Cette fonction recourt à un interpréteur *bash* vulnérable à une attaque de type *ShellShock*. |
| **Phase 2** Armement | Humain | Création d'un RAT (*Remote Access Tool*) permettant d'exécuter des actions sur une cible infectée. |
| **Phase 3** Livraison | Humain | En utilisant le *ShellShock*, téléversement de la charge *reverse-shell*, laquelle traverse le *reverse-proxy* car elle ne correspond à aucune signature connue. |
| **Phase 4** Exploitation | Humain | Utilisation du *reverse-shell* pour altérer la page principale du site de e-commerce afin qu'un employé visitant ladite page télécharge malencontreusement la charge de la phase suivante. |
| **Phase 5** Installation | Humain | Un employé a malencontreusement téléchargé le RAT, lequel s'exécute sur le PC de l'employé puis télécharge l'IA adversaire et l'exécute. |
| **Phase 6** Commande & Contrôle | Humain | **Phase 6a** : téléchargement des outils d'attaque et du logiciel implémentant l'IA adversaire. |
| | IA | **Phase 6b** : l'IA adversaire explore le réseau interne (scan de ports) et découvre le serveur de fichiers hébergeant des répertoires partagés. |
| **Phase 7** Actions sur objectifs | Humain | **Phase 7a** : exfiltration des données sensibles depuis les répertoires partagés. |
| | | **Phase 7b** : envoi des données sensibles vers le serveur CnC (Commande & Contrôle). |
| | | **Phase 7c** : effacement des traces. |

À ce stade de nos travaux, seules les actions de scan de ports mises en œuvre lors des phases 1 et 6 (Table 2 : 1a et 6b) sont automatisées : l'IA adversaire, s'appuyant sur une stratégie de découverte, offre un bon compromis entre efficacité et furtivité qu'un humain ne pourrait vraisemblablement pas améliorer.

**Fig. 6.** — Diagramme de séquence de l'attaque. Les numéros des actions correspondent aux phases de la *Cyber Kill Chain*.

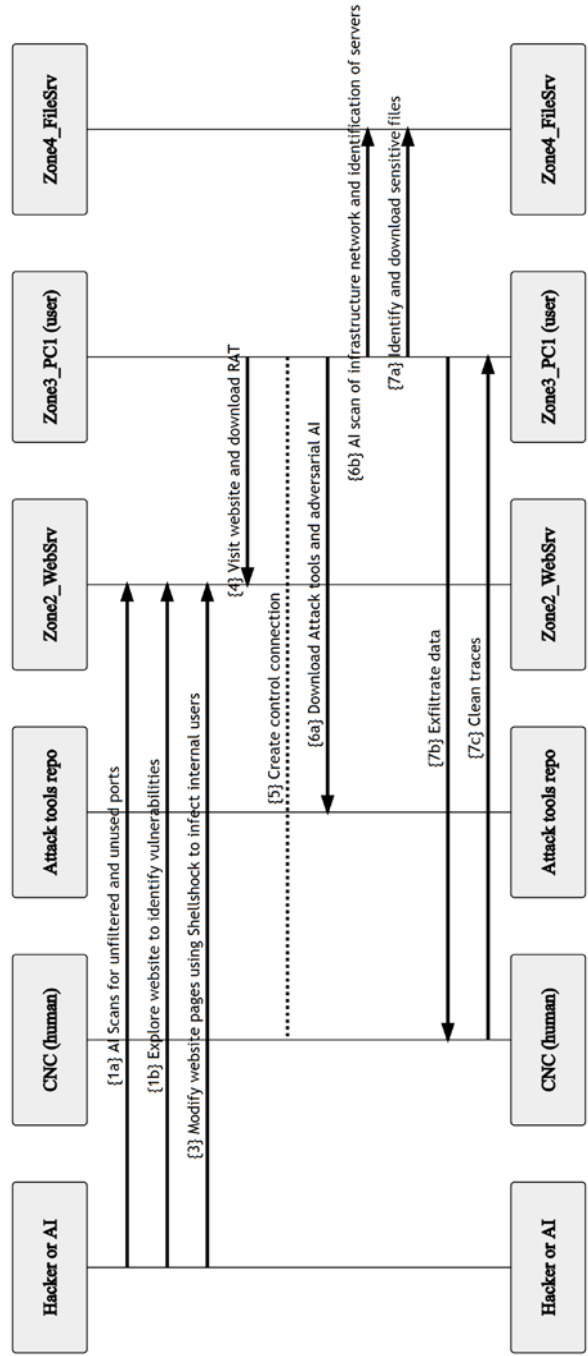

**Fig. 7.** — Cinématique des flux de l'attaque dans le SI. Les numéros des actions correspondent aux phases de la *Cyber Kill Chain*.

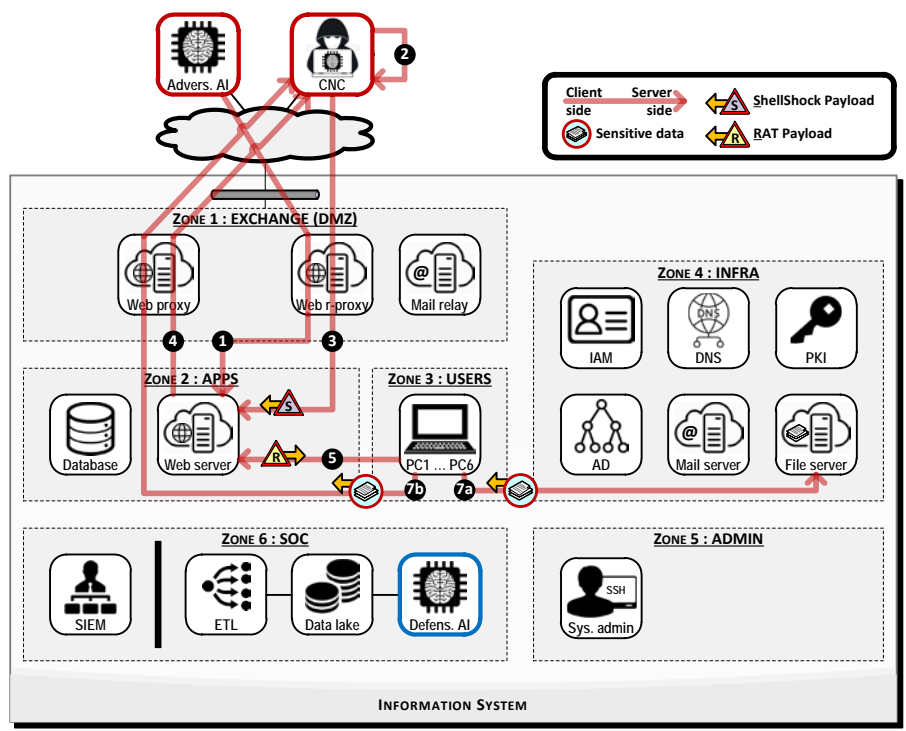

## 5.3 Stratégie adversaire

Notre IA adversaire exécute une centaine d'actions dans un environnement connu et maitrisé. Afin de préparer une attaque à haute vélocité, chaque action est scorée en fonction des découvertes qui en résultent, et, en particulier, l'attente (choix de ne rien faire) est pénalisée. L'exploration est favorisée en privilégiant les séquences courtes et une forte variation de cibles. Quant à l'analyse en détail des cibles, elle est accomplie par des séquences spécifiques d'actions : les informations résultantes sont favorisées au détriment de celles obtenues aisément. Finalement, une stratégie visant à maximiser la discrétion est plus complexe à élaborer. Elle se base sur des hypothèses heuristiques visant à augmenter la difficulté d'être détectée, par exemple en récompensant des méthodes passives d'exploration reposant sur une fréquence d'action réduite.

Le mécanisme actuel de récompense favorise au contraire une stratégie d'analyse davantage focalisée sur la précision des résultats. Pour toutes les cibles, l'agent adversaire identifie l'état des ports réseau (ouvert, fermé, filtré) et les services à l'écoute. En dépit du nombre réduit de types d'actions possibles et de vulnérabilités identifiables, la stratégie de l'IA adversaire est modulable par le mécanisme des récom-penses. L'apprentissage mis en œuvre permet d'atteindre un score nettement supérieur à celui d'un agent aléatoire (sur 100 actions, score d'environ 140 en moyenne pour un agent aléatoire versus 260 pour un agent entraîné), ce qui conforte la pertinence de l'approche. Cependant, nous n'avons pas effectué la comparaison avec un humain.

Dans le cadre de l'apprentissage joint, l'IA adversaire accède au score d'anomalie calculé par l'IA de défense. En incluant la minimisation de ce score dans ses objectifs d'apprentissage, l'IA d'attaque choisit les actions qui semblent les moins anormales lors de sa phase exploratoire. Une fois les capacités de l'IA adversaire étendues à d'autre étapes de la *Cyber Kill Chain*, il deviendra envisageable de leurrer le système de détection en exploitant ses angles morts, permettant à terme de corriger ceux-ci.

### 5.4 Volet défensif

Nous avons implémenté partiellement l'écosystème défensif présenté en Fig. 1 en priorisant la facilité de configuration du procédé de détection. À ce jour, l'analyste humain reste en charge de déterminer les types de données intéressants (ex. : chemin d'un exécutable, succès ou non d'une tentative de connexion) et les indications qui en facilitent le prétraitement (ex. : schéma récurrent dans les noms de fichiers temporaires, noms d'applications générant beaucoup d'événements sans réel intérêt pour la détection).

À terme, cette étape de configuration sera accélérée par recours à des techniques de visualisation et de fouille de données plus avancées.

### 5.5 Résultats

Les résultats obtenus par notre système de reconstruction des graphes du scénario d'attaque sont conformes à nos attentes. Le regroupement des structures fréquentes est rapide (50.000 événements/s sur un processeur standard) et réduit drastiquement (deux ordres de grandeur) la dimension du graphe sans pour autant perdre d'information, préservant ainsi la possibilité d'une fouille des données manuelle lors de l'annotation ultérieure par l'analyste.

Notre système rend donc possible l'utilisation de techniques poussées d'analyse de chacune des connexions du graphe, qui seraient trop gourmandes en temps de calcul sur le graphe original. Cette analyse approfondie permet d'éliminer les chemins rares à l'échelle du système, lequel enregistre des centaines de milliers d'évènements chaque seconde, mais néanmoins régulièrement observés à l'échelle humaine.

Une fois ce nettoyage effectué, parmi les deux millions d'événements présents à l'origine dans nos données d'évaluation, un sous graphe contenant quelques dizaines de nœuds se détache particulièrement du reste. On retrouve dans ce graphe en Fig. 8 tous les événements collectés qui sont liés à l'attaque, après élagage des moins significatifs.

La Fig. 9 représente les événements majeurs de l'attaque par période de temps. Cette représentation permet d'identifier rapidement l'adresse IP de l'attaquant (45.251.23.2) au moment de sa connexion au site web de l'entreprise, l'adresse de son serveur de contrôle (52.95.245.2), ainsi que la machine infectée (PC01) et les commandes les plus marquantes effectuées sur celle-ci (lancement du RAT, script python effectuant le scan de port, récupération du fichier et son export, et suppression des traces du passage de l'attaquant qui tente ainsi de dissimuler son action).

Ces deux visualisations sont complémentaires. La ligne temporelle des évènements majeurs (Fig. 9) offre à l'analyste la possibilité d'évaluer rapidement la pertinence de l'alerte, tandis que le graphe élagué d'intérêt majeur (Fig. 8) est dévolu à une analyse plus poussée car il procure les détails nécessaires permettant d'accélérer la remédiation de l'incident de sécurité.

Les premiers résultats de notre système de rétroaction sont positifs, notamment en ce qui concerne sa capacité d'apprentissage incrémental des réseaux de neurone auto-encodeurs et sa vitesse d'entrainement du procédé de reconstruction. Néanmoins, notre système de rétroaction n'étant pas finalisé, il ne peut donc être évalué exhaustivement. Ceci fera l'objet de futurs travaux dès 2020.

**Fig. 8.** — Graphe élagué d'intérêt majeur. Les nœuds en violet représentent les commandes exécutées, ceux en vert les événements réseau. La flèche du temps s'oriente de gauche à droite.

**Fig. 9.** — Ligne temporelle des événements majeurs.

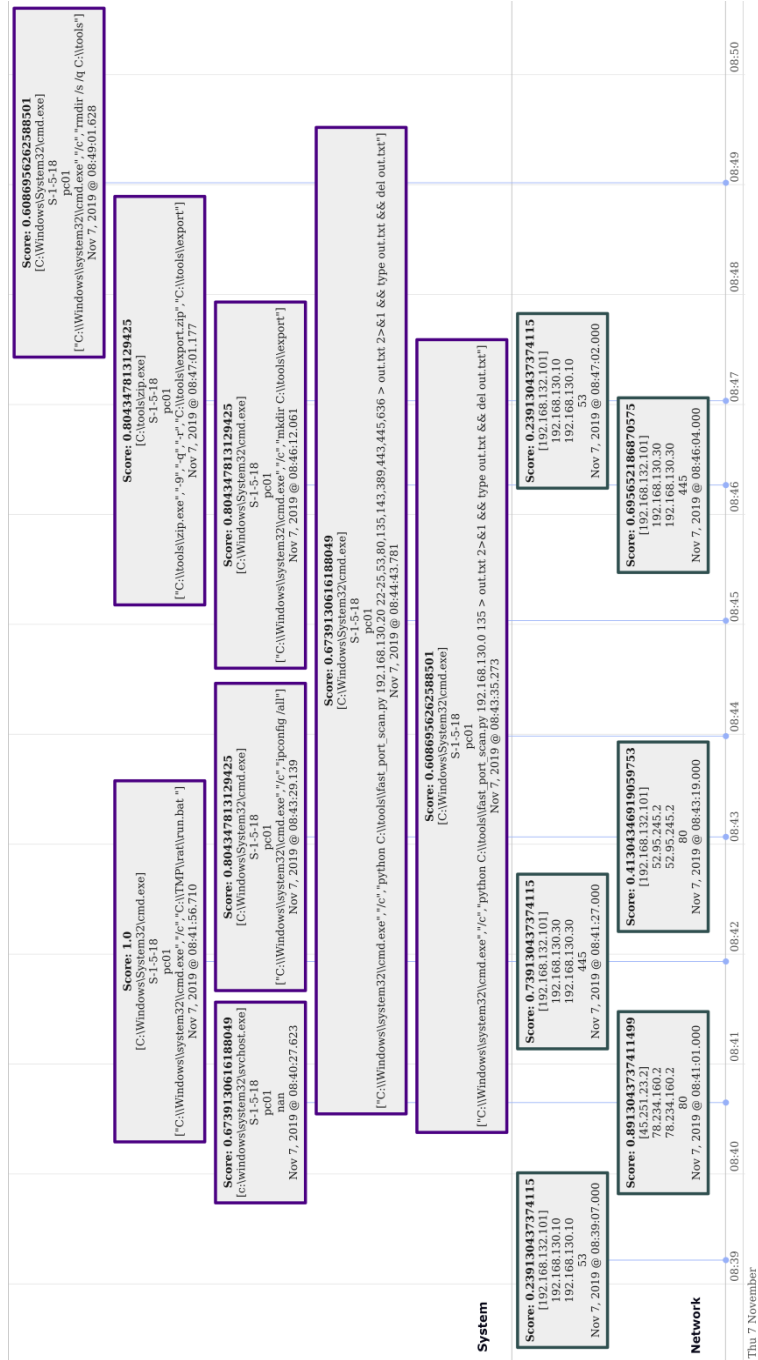

# 6 Conclusion

Notre procédé est conçu pour analyser d'importants volumes d'événements de sécurité, au fil de l'eau, en profondeur et sans règles prédéfinies, en évaluant le caractère inhabituel de groupes d'évènements corrélés, afin de produire une alerte explicable prenant la forme d'un graphe reconstruisant un scénario d'attaque.

Un tel procédé défensif typé UEBA étant naturellement sensible au leurrage, nous avons conçu un écosystème dans lequel un procédé d'IA adversaire s'oppose à un procédé d'IA défensive, dans une approche dérivée de l'apprentissage joint.

Le système d'attaque identifie rapidement les vulnérabilités de la défense facilitant ainsi leur correction. La création de cet écosystème est d'autant plus complexe que nous développons conjointement les deux systèmes IA.

L'objectif poursuivi à moyen terme est triple : d'une part, permettre à chaque IA de contribuer à améliorer l'autre, d'autre part, de pallier au moins en partie la difficulté de collecter des données d'attaque représentatives de notre modèle de menace, et enfin, de vérifier et qualifier le fonctionnement de notre système défensif afin d'étendre notre méthodologie à d'autres solutions ou produits.

En particulier, ce dernier apport consistera à identifier les faiblesses d'un système défensif, car si le succès d'une attaque adversaire démontre logiquement une faiblesse de défense, a contrario l'échec d'une attaque ne prouve rien puisque il peut être imputé aux qualités du défenseur autant qu'à l'insuffisance de l'adversaire.

Nos travaux futurs nous amèneront à nous pencher sur l'épineuse problématique de la qualification et la certification de l'IA adversaire si cette dernière doit elle-même être utilisée pour qualifier voire certifier des procédés défensifs.

# 7 Références